\documentclass{article} 
\usepackage[final]{colm2025_conference}

\usepackage{microtype}
\usepackage{hyperref}
\usepackage{url}
\usepackage{booktabs}
\usepackage{xcolor}
\usepackage{lineno}
\usepackage{amsmath}
\usepackage[whole]{bxcjkjatype}
\usepackage[normalem]{ulem}
\useunder{\uline}{\ul}{}
\usepackage{multirow} 
\usepackage{graphics}
\usepackage[pdftex]{graphicx}

\definecolor{darkblue}{rgb}{0, 0, 0.5}
\hypersetup{colorlinks=true, citecolor=darkblue, linkcolor=darkblue, urlcolor=darkblue}

\title{Efficient Construction of Model Family through \\Progressive Training Using Model Expansion}

\author{Kazuki Yano${}^{\dag}$, Sho Takase${}^{\dag, \ddag}$, Sosuke Kobayashi${}^{\dag}$, Shun Kiyono${}^{\ddag}$, Jun Suzuki${}^{\dag}$.  \\
${}^{\dag}$Tohoku University\\
${}^{\ddag}$SB Intuitions \\
\texttt{yano.kazuki@dc.tohoku.ac.jp} \\
}

\newcommand{\independent}{\texttt{Independent}}
\newcommand{\progressive}{\texttt{Progressive}}

\begin{document}

\ifcolmsubmission
\linenumbers
\fi

\maketitle

\begin{abstract}
As Large Language Models (LLMs) gain widespread practical application, offering model families with varying parameter sizes has become standard practice to accommodate diverse computational requirements.
Traditionally, each model in the family is trained independently, incurring computational costs that scale additively with the number of models. 
In this work, we propose an efficient method for constructing model families via progressive training, where smaller models are incrementally expanded to larger sizes to create a complete model family.
Through extensive experiments on a model family ranging from 1B to 8B parameters, we show that our approach reduces total computational cost by approximately 25\% while maintaining comparable performance to independently trained models.
Moreover, by strategically adjusting the maximum learning rate based on model size, our method outperforms the independent training across various metrics.
Beyond these improvements, our approach also fosters greater consistency in behavior across model sizes.
\end{abstract}

\section{Introduction}
As Large Language Models (LLMs) gain widespread practical application, providing models with (i) a consistent architecture and (ii) varying parameter sizes (hereafter referred to as a \textbf{model family}) has become standard practice in the NLP community. 
For instance, Llama 3.1 includes models with 8B, 70B, and 405B parameters~\citep{grattafiori2024llama3herdmodels}, while Gemma~3 provides 1B, 4B, 12B, and 27B parameter variants~\citep{team2025gemma3}
.
Similarly, Qwen2.5 offers a model family with 0.5B, 1.5B, 3B, 7B, 14B, 32B, and 72B parameters~\citep{yang2024qwen2.5}. 

\begin{figure}[t]
    \centering \includegraphics[width=0.8\columnwidth]{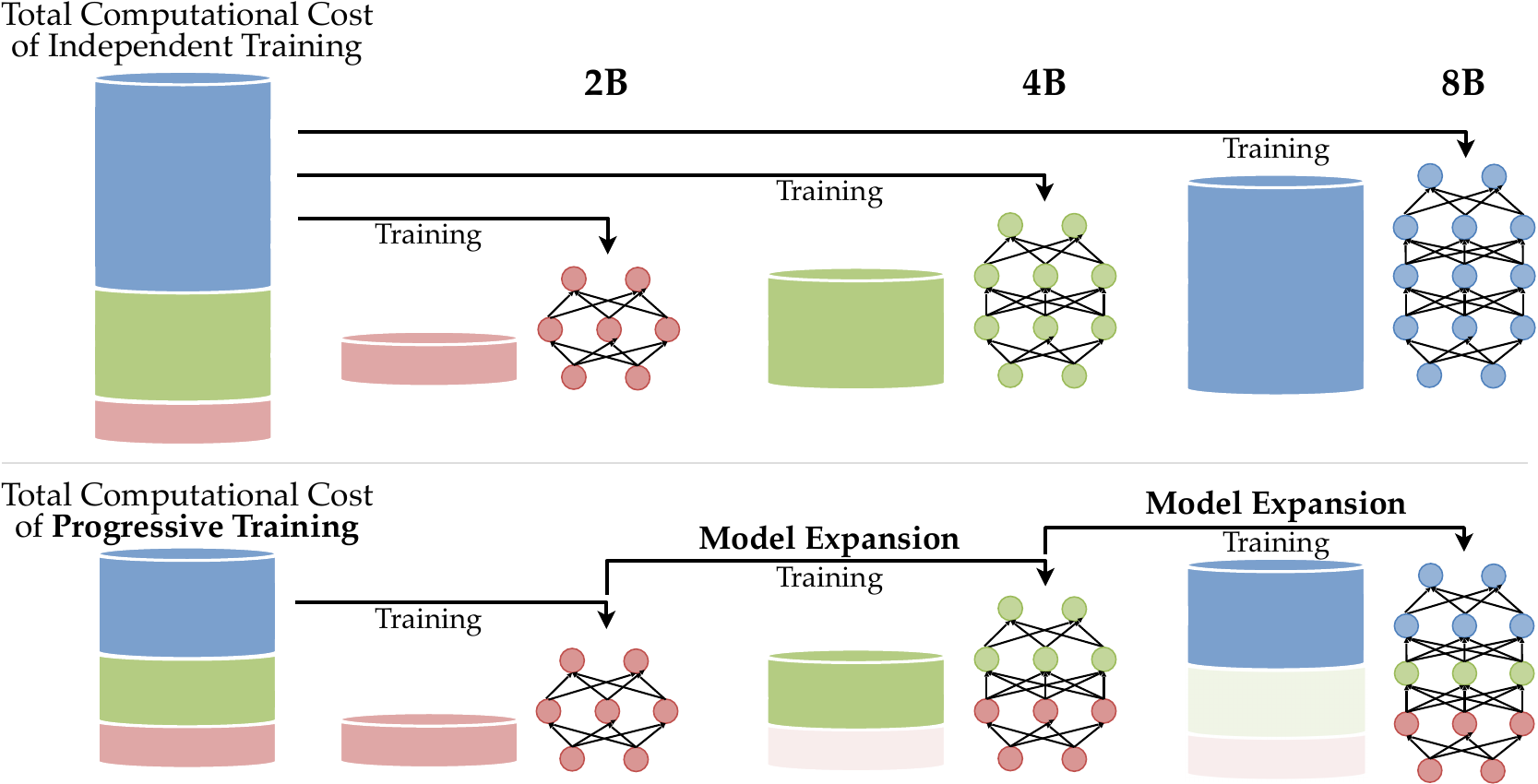}
    \caption{Comparison of approaches for constructing a model family. (\textit{Top}): Conventional approach, where each model in the family (2B, 4B, 8B) is trained independently from scratch. The total computational cost is the sum of the training costs for all individual models. (\textit{Bottom}): 
    Proposed progressive training utilizes model expansion, where smaller models are expanded to initialize larger ones. The total cost equals only that of the largest model (8B).}
    \label{fig:upsycle_concept} 
\end{figure}

Such model families are designed to address a wide range of computational constraints and application scenarios.
Smaller models offer faster inference and lower resource consumption, making them suitable for daily tasks and deployment in resource-constrained environments such as smartphones and edge devices~\citep{abdin2024phi}. 
In contrast, larger models are deployed for scenarios requiring advanced reasoning capabilities and complex task processing, typically on large-scale servers~\citep{wei2022emergent}.
The standard approach to constructing a model family involves training each model independently from scratch.
However, training large-scale models demands extensive resources, e.g., thousands of GPU days~\citep{ touvron2023llama}.
The total computational cost of constructing a model family poses a significant burden on its builders. (Figure~\ref{fig:upsycle_concept} (\textit{Top})).
This motivates us to explore more efficient methods for model family construction.

We identify \textbf{model expansion} as a potential approach to more efficiently constructing model families, including large-scale models.
Model expansion leverages the parameters of pre-trained smaller models as initialization for training larger models~\citep{chen-etal-2022-bert2bert,du2024stacking}.
However, prior work on model expansion  has primarily aimed at producing a single final model, with limited focus on the potential utility of intermediate models.

In this work, we empirically explore model expansion as a means to efficiently construct a model family.
Specifically, we propose a method that repeatedly applies model expansion to construct models from smaller to larger sizes, namely \textbf{progressive training}, thereby reducing the total training cost of constructing a model family (Figure~\ref{fig:upsycle_concept} (\textit{Bottom})).

Through our experiments with a model family ranging from 1B to 8B parameters, we demonstrate that the proposed method reduces the total computational cost of constructing a complete model family by approximately 25\% compared to training each model independently\footnote{
This 25\% reduction corresponds to savings of approximately 3.2K GPU hours in our experimental setup.
}.
Moreover, by adjusting the maximum learning rate based on model size, our method consistently achieves superior performance across a range of benchmark tasks relative to independently trained models. 
We also found that progressive training yields greater behavioral consistency across the model family, as indicated by lower Kullback-Leibler (KL) divergence between models' output distributions.

\section{Task Definition and Notation Rules}\label{sec:definition}
We consider a model family consisting of models with monotonically increasing sizes and consistent architecture.
Formally, we define this model family as a sequence of models of sizes $[X_1, X_2, \ldots, X_n]$ and corresponding sequence of model parameters $[\theta_1, \theta_2, \ldots, \theta_n]$, where $\theta_i \in \mathbb{R}^{X_i}$ denotes the parameters of the $i$-th model.
For each model of size $X_i$, let $T_i$ represent the number of training tokens used to train the corresponding model.
The computational cost (FLOPs) required to train a given model is determined by the model size $X_i$ and the number of tokens $T_i$.
Following \cite{brown2020gpt3}, we define FLOPs as: $\mathrm{FLOPs}(X_i, T_i) = 6X_iT_i$.
Note that our analysis does not depend on this specific formulation and can be generalized to other reasonable cost approximations~\citep{hoffmann2022chinchilla}.

Let $T_i^{\textrm{scratch}}$ denote the number of tokens for training a randomly initialized model of size $X_i$ from scratch.
The total computational cost of constructing the entire model family is then given by $\sum_{i=1}^{n}\mathrm{FLOPs}(X_{i}, T_{i}^{\textrm{scratch}})$.
The task of this work is to reduce the overall computational cost while preserving the performance of each model.
Formally, we aim to construct each model under the following constraint:
\begin{align}
    \sum_{i=1}^{n}\mathrm{FLOPs}(X_{i}, T_{i}^{\textrm{scratch}}) > \sum_{i=1}^{n}\mathrm{FLOPs}(X_{i}, T_{i}^{\textrm{prog}}),
    \label{eq:restriction}
\end{align}
where $T_i^{\mathrm{prog}}$ denotes the number of tokens used to train the model of size $X_i$ under the progressive training approach (Section~\ref{sec:method}).

\section{Method: Progressive Training}
\label{sec:method}
We propose a method for constructing language models of varying sizes more efficiently than the common practice, i.e., training each model independently from scratch.
In the common approach, the cost of training a model of size $X_{i}$ from scratch using $T_{i}^{\textrm{scratch}}$ tokens\footnote{
Note that the choice of $T_i^{\textrm{scratch}}$ is arbitrary depending on available resources, though most researchers follow the Chinchilla law~\citep{hoffmann2022chinchilla} as a principled approach to determine the compute-optimal number of training tokens.
Our experimental setup also adopts the Chinchilla law (Section~\ref{subsec:data_size}).
} is $\mathrm{FLOPs}(X_{i}, T_{i}^{\textrm{scratch}})$, leading to a total cost of $\sum_{i=1}^{n}\mathrm{FLOPs}(X_{i}, T_{i}^{\textrm{scratch}})$ for the entire model family.
A simple strategy to satisfy Equation (\ref{eq:restriction}) is to reduce the number of training tokens for each model by selecting $T_{i}^{\textrm{prog}}$ such that $T_{i}^{\textrm{scratch}} > T_{i}^{\textrm{prog}}$ for $i > 1$.
However, naively reducing the number of training tokens typically results in degraded performance if other factors remain unchanged.
Instead, we initialize the $i$-th model of size $X_{i}$ for $i>1$ by using a better initialization than random, which is obtained by applying model expansion to the previously trained model sizesize $X_{i-1}$.
We refer to this process as \textbf{progressive training}.

Specifically, our progressive training begins by training an initial model of size $X_1$ using $T_{1}^{\textrm{scratch}}$ tokens.
At each subsequent stage, we initialize the next model using a model expansion method $f$,
\begin{equation}
\theta_{i+1}^{\textrm{init}} = f(\theta_i; X_{i+1}) \quad (i \geq 1),
\end{equation}

where $f(\cdot; X_{i+1}): \mathbb{R}^{X_i} \to \mathbb{R}^{X_{i+1}}$ is an off-the-shelf model expansion method.
The expanded parameters $\theta_{i+1}^{\textrm{init}}$ serve as an effective initialization for training the subsequent model.

There exists a wide range of possible token allocation patterns $[T_{1}^{\textrm{prog}}, \  T_{2}^{\textrm{prog}}, ..., T_{n}^{\textrm{prog}}]$ that satisfy the constraint in Equation (\ref{eq:restriction})'s condition regarding the total computational cost.
In this work, we determine each $T_{i+1}^{\textrm{prog}}$ such that the overall computational cost matches that of training the largest model $X_{n}$ from scratch:
\begin{align}
    \mathrm{FLOPs}(X_{n}, T_{n}^{\textrm{scratch}}) = \sum_{i=1}^{n}\mathrm{FLOPs}(X_{i}, T_{i}^{\textrm{prog}}).
    \label{eq:this_work}
\end{align}
Specifically, we set $T_{i+1}^{\textrm{prog}}$ to satisfy the following FLOPs constraint:
\begin{equation}\label{eq:prog_flops}
    \mathrm{FLOPs}(X_{i+1}, T_{i+1}^{\textrm{prog}} )= \mathrm{FLOPs}(X_{i+1}, T^{\textrm{scratch}}_{i+1}) - \sum_{j=1}^i \mathrm{FLOPs}(X_j, T^{\textrm{prog}}_j).
\end{equation}
Intuitively, we allocate the computational cost of training a model of size $X_{i+1}$ from scratch, subtracting the computational cost already used in previous stages.
Through this procedure, we can obtain an entire model family from size $X_1$ to $X_{n}$ with a total computational cost equal $\mathrm{FLOPs}(X_{n}, T_{n}^{\textrm{scratch}})$. 

\paragraph{How we choose $f$.} In this work, we adopt bert2BERT~\citep{chen-etal-2022-bert2bert} as our model expansion method for the following reason.
Unlike approaches that focus solely on depth expansion (such as stacking~\citep{ du2024stacking}), bert2BERT allows us to increase both the width and depth dimensions of Transformer models~\citep{vaswani2017attention}, offering greater flexibility\footnote{
Although recent methods such as LEMON~\citep{wang2024lemon} have emerged, LEMON builds upon the foundation of bert2BERT with only minor modifications to its core mechanisms. We chose bert2BERT for its simplicity and ease of implementation.
}.
Specifically, for width expansion,  bert2BERT increases the hidden dimensions by duplicating the weights of linear layers; for depth expansion, it stacks additional layers by duplicating pre-trained ones\footnote{The original bert2BERT paper~\citep{chen-etal-2022-bert2bert} introduces two expansion variants: AKI and FPI. In this work, we exclusively adopt the AKI}.
This flexibility to expand along both dimensions is particularly valuable as we repeatedly apply model expansion in progressive training.

\section{Experiment}
\label{sec:exp_pretraining}
To evaluate the effectiveness of progressive training, we pre-train and compare two model families: one with independently training each model from scratch (\independent{}) and the other with progressive training (\progressive{}).
Specifically, we demonstrate that progressive training improves computational efficiency over independent training, and the performance of the two model families is comparable.

\subsection{Experimental Setup}
\label{subsec:experimental_setup}

We used FineWeb-Edu~\citep{penedo2024fineweb} as the training data for pre-training and adopted the GPT-2~\citep{radford2019language} tokenizer for tokenization.
While progressive training can accommodate any parameter size increase, in this work, we adopted a configuration where the parameter count doubles at each stage.
Specifically, we construct a model family with sizes $[X_1=1\textrm{B}, X_2=2\textrm{B}, X_3=4\textrm{B}, X_4=8\textrm{B}]$\footnote{
The specific width and depth settings for each model are detailed in Appendix~\ref{appendix:exp_config}.
}.
All models follow the Llama architecture~\citep{touvron2023llama} with a maximum input sequence of 1024 tokens.
Each model was trained using a cosine learning rate scheduler, with a maximum learning rate of $3.0 \times 10^{-4}$.

To evaluate each pre-trained model in the family, we measured perplexity on the validation data (Valid) from FineWeb-Edu and WikiText~\citep{merity2017pointer}. 
To more comprehensively assess the performance of pre-trained models, we also evaluated zero-shot performance across multiple downstream tasks.
Specifically, we include tasks including language modeling (LAMBADA~\citep{paperno-etal-2016-lambada}), 
commonsense reasoning (WinoGrande~\citep{sakaguchi2021winogrande}, PIQA~\citep{piqa}, HellaSwag~\citep{zellers-etal-2019-hellaswag}), 
and question answering (ARC-e, ARC-c~\citep{clark2018think}, OBQA~\citep{mihaylov-etal-2018-suit}).

\subsection{Training Data Size and Computational Cost}\label{subsec:data_size}
We prepare the following two data sizes, (i) Chinchilla law~\citep{hoffmann2022chinchilla} and (ii) 2x Chinchilla law, to determine the number of training tokens $T_{i}^{\textrm{scratch}}$ for each model in a model family.
Chinchilla law provides the optimal FLOPs required to achieve a specific loss.
The guideline is to use 20 tokens per model parameter for training~\citep{hoffmann2022chinchilla}.
2x Chinchilla law is used to simulate the case such that the amount of training data greatly exceeds the optimal values indicated by the Chinchilla law.
In fact, exceeding the Chinchilla law has become a standard practice in recent LLM literature~\citep{sardana2024beyond}.

In the progressive training approach, we determine the number of training tokens $T_{i}^{\textrm{prog}}$ for each model to satisfy the constraint defined in Equation (\ref{eq:prog_flops}), ensuring it matches the total computational cost equals that of training the largest model from scratch.
Under the 2x Chinchilla law setting, this constraint yields the following token allocations: $T_1^{\textrm{prog}} = 40\textrm{B}, T_2^{\textrm{prog}} = 60\textrm{B}, T_3^{\textrm{prog}} = 120\textrm{B}, T_4^{\textrm{prog}} = 240\textrm{B}$\footnote{
For the Chinchilla law setting, $T_i^{\textrm{prog}}$ is shown in Appendix~\ref{appendix:chinchilla_setting}.}. 

One of the key advantages of progressive training is its ability to obtain models of multiple sizes efficiently.
As defined in Section~\ref{sec:definition}, the computational cost is calculated as $\textrm{FLOPs} = 6XT$, where $X$ denotes the number of model parameters and $T$ is the number of training tokens.
For instance, independently training each model in the family $[1\mathrm{B}, 2\mathrm{B}, 4\mathrm{B}, 8\mathrm{B}]$ under the 2x Chinchilla law would require FLOPs of $[0.24\textrm{Z}, 0.96\textrm{Z}, 3.84\textrm{Z}, 15.4\textrm{Z}]$ respectively. 
Thus, constructing the entire model family would require $20.4\mathrm{ZFLOPs}$ in total.

In contrast, with progressive training, the total computational cost required to construct the entire model family is equivalent to the computational cost required to train the largest-size, i.e., 8B, model. 
Therefore, the computational cost is $15.4\mathrm{ZFLOPs}$.
Compared to training each model independently, progressive training can reduce computational cost by approximately 25\%. 
This demonstrates that progressive training improves the computational efficiency of constructing a model family.

\subsection{Results}

\begin{table}[t]
    \centering
    \small
    \tabcolsep 1.5pt
    \begin{tabular}{cc|ccccccccc}
        \toprule 
         & \multirow{2}{*}{} 
            & \multicolumn{2}{c}{Perplexity \textdownarrow}  
            & \multicolumn{7}{c}{Accuracy \textuparrow}         \\ 
        \cmidrule(r){3-4} \cmidrule(r){5-11} 
            & 
            & Valid & Wikitext  
            & LAMBADA  & ARC-e  & ARC-c  & Winogrande & PIQA  & OBQA & HellaSwag  \\ 
        \midrule
        \multirow{1}{*}{1B} 
            & \independent{}
            & 13.14  & 22.81
            & 39.3  & 59.3  & 31.7  & 54.0  & 70.5  & 35.6  & 46.9   \\
        \midrule
        \multirow{2}{*}{2B} 
            & \independent{}
            & 11.30  & \textbf{18.57}
            & \textbf{45.5}  & \textbf{65.2}  & \textbf{37.7}  & 55.3  & 72.1  & 38.8  & 54.3   \\
            & \progressive{}
            & \textbf{11.29}  & 18.74
            & 45.1  & 63.6 & 36.6  & \textbf{57.4}  & \textbf{72.7}  & \textbf{39.2}  & \textbf{54.7}  \\
        \midrule
        \multirow{2}{*}{4B} 
            & \independent{}
            & 9.91  & 15.46
            & 49.5  & 68.8  & \textbf{41.1}  & 57.2  & 75.1  & \textbf{43.8}  & 60.5   \\
            & \progressive{}
            & \textbf{9.87}  & \textbf{15.12}
            & \textbf{51.0}  & \textbf{69.4}  & 40.0  & \textbf{58.6}  & 75.1  & 40.6  & \textbf{61.2} \\
        \midrule
        \multirow{2}{*}{8B} 
            & \independent{}
            & 8.65  & 12.24
            & 53.9  & 71.8  & 43.1  & \textbf{62.4}  & 76.2  & 42.6  & 65.8 \\
            & \progressive{}
            & \textbf{8.61}  & \textbf{11.98}
            & \textbf{55.4}  & \textbf{73.5}  & \textbf{45.1}  & 62.1  & \textbf{76.6}  & \textbf{45.6} & \textbf{67.5}  \\
        \bottomrule
    \end{tabular}
    \caption{
    Evaluation results of pre-trained models under the Chinchilla law setting. For each model size, we compare the performance of models trained independently from scratch (\independent{}) versus those built using progressive training (\progressive{}). 
    }
    \label{tab:chinchilla_lr_fixed}
\end{table}
\begin{table}[t]
    \centering
    \small
    \tabcolsep 1.5pt
    \begin{tabular}{cc|ccccccccc}
        \toprule 
         & \multirow{2}{*}{} 
            & \multicolumn{2}{c}{Perplexity \textdownarrow}  
            & \multicolumn{7}{c}{Accuracy \textuparrow}         \\ 
        \cmidrule(r){3-4} \cmidrule(r){5-11} 
            & 
            & Valid & Wikitext  
            & LAMBADA  & ARC-e  & ARC-c  & Winogrande & PIQA  & OBQA & HellaSwag  \\ 
        \midrule
        \multirow{1}{*}{1B} 
            & \independent{}
            & 12.04  & 20.56
            & 42.0  & 62.3  & 34.6  & 55.5  & 72.0  & 35.0  & 51.6   \\
        \midrule
        \multirow{2}{*}{2B} 
            & \independent{}
            & 10.44  & 17.11
            & 48.5  & 66.9  & \textbf{38.4}  & 56.6  & \textbf{75.3}  & \textbf{41.6}  & 58.0   \\
            & \progressive{}
            & \textbf{10.43}  & \textbf{16.94}
            & \textbf{50.3}  & \textbf{70.0} & 38.1  & \textbf{58.5}  & 73.8  & 41.4  & \textbf{59.0}  \\
        \midrule
        \multirow{2}{*}{4B} 
            & \independent{}
            & 9.18  & 13.99
            & 51.7  & 71.7  & 43.2  & 59.2  & \textbf{76.7}  & 40.8  & 63.3   \\
            & \progressive{}
            & \textbf{9.00}  & \textbf{13.63}
            & \textbf{51.9}  & \textbf{72.6}  & \textbf{43.9}  & \textbf{61.0}  & 76.1  & \textbf{43.0}  & \textbf{65.6} \\
        \midrule
        \multirow{3}{*}{8B} 
            & \independent{}
            & 7.85  & 10.23
            & 55.2  & 74.5  & 47.4  & 62.4  & 77.0  & \textbf{46.4}  & 67.9 \\
            & \progressive{}
            & \textbf{7.74}  & \textbf{10.21}
            & \textbf{57.8}  & \textbf{75.3}  & 46.7  & \textbf{64.6}  & 77.3  & 44.6 & \textbf{70.2}  \\
            &  \hspace{3pt} \texttt{+Fixed Data}
            & \textbf{7.73} & \textbf{10.22} 
            & 56.2   & 74.8  & \textbf{47.8}  & 63.1  & \textbf{77.8}  & 45.2 & \textbf{70.4} \\
        \bottomrule
    \end{tabular}
    \caption{
    Evaluation results of pre-trained models under the 2x Chinchilla law setting. Performance is compared between models trained independently from scratch (\independent{}) and those built using progressive training (\progressive{}). \progressive{}\texttt{+Fixed Data} indicates training on a fixed dataset of 320B tokens.
    }
    \label{tab:2x_chinchilla_lr_fixed}
\end{table}
\begin{figure}[t]
    \centering
    \includegraphics[width=\linewidth]{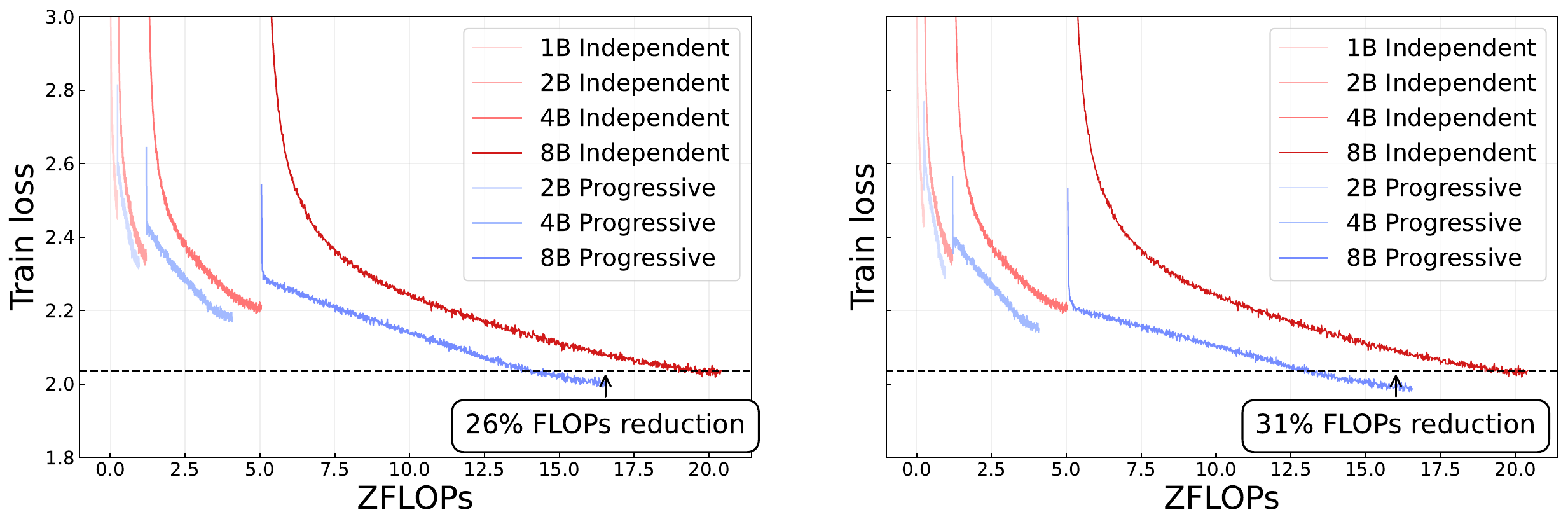}
    \caption{Train loss curves comparing models trained by \texttt{Independent} versus \texttt{Progressive} approach with the 2x Chinchilla law setting. (\textit{Left}): Models trained with fixed maximum learning rate, achieving 26\% FLOPs reduction. (\textit{Right}): Models trained with maximum learning rate adjustment, from $1.5 \times 10^{-3}$ (1B) to $3.0 \times 10^{-4}$ (8B), achieving 31\% FLOPs reduction.}
    \label{fig:loss_curve} 
\end{figure}
\begin{figure}[t]
    \centering
    \includegraphics[width=1\columnwidth]{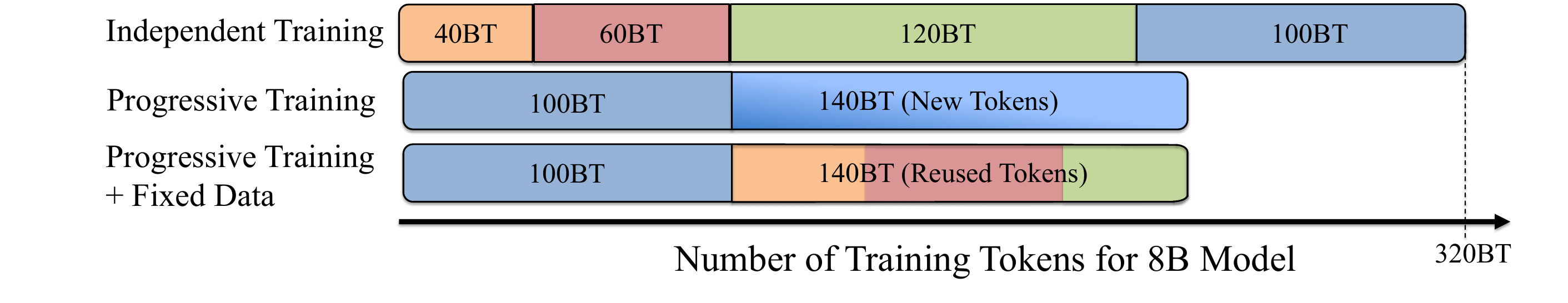}
    \caption{Comparison of token allocation for the 8B model training. (\textit{Top}): \independent{} uses a standard approach with 320B tokens. (\textit{Middle}): \progressive{} uses 240B tokens (100B + new 140B tokens). (\textit{Bottom}): \progressive{}\texttt{+Fixed Data} maintains the same 240B total tokens as \progressive{} by reusing 140B tokens from previous stages.}
    \label{fig:fixed_data} 
\end{figure}
\begin{table}[t]
    \centering
    \tabcolsep 1.5pt
    \small
    \begin{tabular}{cc|ccccccccc}
        \toprule 
         & \multirow{2}{*}{} 
            & \multicolumn{2}{c}{Perplexity \textdownarrow}  
            & \multicolumn{7}{c}{Accuracy \textuparrow}         \\ 
        \cmidrule(r){3-4} \cmidrule(r){5-11} 
            & 
            & Valid & Wikitext  
            & LAMBADA  & ARC-e  & ARC-c  & Winogrande & PIQA  & OBQA & HellaSwag  \\ 
        \midrule
        \multirow{1}{*}{1B} 
            & \independent{}
            & 12.04  & 20.56
            & 42.0  & 62.3  & 34.6  & 55.5  & 72.0  & 35.0  & 51.6   \\
        \midrule
        \multirow{2}{*}{2B} 
            & \independent{}
            & 10.44  & 17.11
            & 48.5  & 66.9  & 38.4  & 56.6  & 75.3  & 41.6  & 58.0   \\
            & \progressive{}
            & \textbf{10.07}  & \textbf{16.17}
            & \textbf{52.6}  & \textbf{71.8} & \textbf{42.2}  & \textbf{61.9}  & \textbf{75.0}  & \textbf{40.8}  & \textbf{62.5}  \\
        \midrule
        \multirow{2}{*}{4B} 
            & \independent{}
            & 9.18  & 13.99
            & 51.7  & 71.7  & 43.2  & 59.2  & 76.7  & 40.8  & 63.3   \\
            & \progressive{}
            & \textbf{8.72}  & \textbf{13.13}
            & \textbf{55.7}  & \textbf{74.5}  & \textbf{47.8}  & \textbf{65.2}  & \textbf{77.4}  & \textbf{45.8}  & \textbf{68.1} \\
        \midrule
        \multirow{3}{*}{8B} 
            & \independent{}
            & 7.85  & 10.23
            & 55.2  & 74.5  & 47.4  & 62.4  & 77.0  & 46.4  & 67.9 \\
            & \progressive{}
            & \textbf{7.64}  & 10.10
            & \textbf{59.0}  & \textbf{76.7}  & 48.3  & \textbf{65.8}  & \textbf{78.3}  & 46.6 & 71.0  \\
            &  \hspace{3pt} \texttt{+Fixed Data}
            & \textbf{7.64} & \textbf{9.77} 
            & 58.0   & 76.1  & \textbf{50.6}  & 64.7  & 78.2  & \textbf{47.2} & \textbf{71.2} \\
        \bottomrule
    \end{tabular}
    \caption{
    Evaluation results of pre-trained models with the maximum learning rate adjustment strategy described in Section~\ref{subsec:learning_rate_adjustment}. \progressive{} models consistently outperform \independent{} models. \texttt{+Fixed Data} indicates training on a fixed dataset of 320B tokens.
    }
\label{tab:2x_chinchilla_lr_strategy}
\end{table}

\paragraph{Model Performance.}
Tables~\ref{tab:chinchilla_lr_fixed} and~\ref{tab:2x_chinchilla_lr_fixed} present the evaluation results of pre-trained models in the Chinchilla law and 2x Chinchilla law settings, respectively. 
Our progressive training approach (\progressive{}) achieves performance comparable to or better than models that are independently trained from scratch (\independent{}) across different parameter sizes in both settings.
Notably, for the largest model (8B), \progressive{} shows improvements in perplexity and most downstream tasks. 

\paragraph{Computational Cost.}
In Figure~\ref{fig:loss_curve} (\textit{Left}), for a given computational budget, i.e., ZFLOPs, \progressive{} consistently outperforms \independent{}.
In addition, \progressive{} reaches the same training loss as \independent{} while reducing the total FLOPS by 26\%.

\paragraph{Does Progressive Training Exploit More Unique Data? No.}

For each model size $X_i (i > 1)$ in our model family, we use fewer tokens in progressive training than independent training ($T_i^{\textrm{prog}} < T_i^{\textrm{scratch}}$). 
However, when considering the entire model family construction process, progressive training consumes more total tokens than training only the largest model from scratch.
As detailed in Section~\ref{subsec:data_size}, the total number of tokens processed in progressive training is $\sum_{i=1}^4{T_i^{\textrm{prog}}} = 460\textrm{B}$ tokens, which exceeds the $T_4^{\textrm{scratch}} = 320\textrm{B}$ tokens used when training the 8B model independently, even though the total computational cost is the same (Figure~\ref{fig:fixed_data}, (\textit{Top}) and (\textit{Middle})).
Thus, it is possible that progressive training benefits unfairly from exposure to a larger volume of unique training data.
To isolate this factor, we introduced a controlled setting denoted as \progressive{}\texttt{+Fixed Data}, in which total amount of unique data is capped at $320\textrm{B}$ tokens -- the same amount used when independently training the 8B model. 
Once the 320B tokens were consumed during the training of the 8B model, the remaining $460-320=140$B tokens are reused, i.e., the second epoch begins, as illustrated in Figure~\ref{fig:fixed_data} (\textit{Bottom}).

The result is shown in Table~\ref{tab:2x_chinchilla_lr_fixed}, in which 8B \progressive{}\texttt{+Fixed Data} achieves comparable performance.
This confirms that the observed improvements from progressive training are not attributable to access to a greater quantity of unique data.

\subsection{Effectiveness of Maximum Learning Rate Adjustment}
\label{subsec:learning_rate_adjustment}
In LLM pre-training, selecting an appropriate learning rate requires a careful balance between optimization efficiency and training stability. 
Higher learning rates can accelerate convergence and potentially lead to better final model performance~\citep{takase2023spike}, but they must be tuned carefully, especially with respect to model size. 
As models scale up, they become increasingly sensitive to the learning rate, typically requiring smaller learning rates to preserve training stability~\citep{touvron2023llama, wortsman2024smallscale}.
For example, in our preliminary experiments, we observed that a learning rate of $1.5 \times 10^{-3}$ enabled effective and stable training for a 1B model. 
However, applying the same learning rate to an 8B model resulted in the loss value spikes that ultimately led to training collapse\footnote{A visualization of this training instability with the 8B model is provided in the Appendix~\ref{appendix:8b_loss}.}.

Progressive training method is characterized by starting training from smaller models. 
Therefore, it is possible to employ high learning rates to enhance performance during small-model training while lowering the learning rate for larger models to stabilize training.
In this experiment, we apply this maximum learning rate adjustment under the 2x Chinchilla law setting, linearly decreasing the maximum learning rate from $1.5 \times 10^{-3}$ for the 1B model to $3.0 \times 10^{-4}$ for the 8B model, in accordance with the increase in model size, while maintaining a cosine decay schedule during training for all models\footnote{The learning rates for each model size are described in Appendix~\ref{appendix:lr_adjustment}.}.

Table~\ref{tab:2x_chinchilla_lr_strategy} presents the results of our maximum learning rate adjustment strategy. 
\progressive{} consistently outperforms the \independent{} counterparts, with more substantial improvements than those observed with no learning rate adjustments (Table~\ref{tab:2x_chinchilla_lr_fixed}). 
As illustrated in Figure~\ref{fig:loss_curve}~(\textit{Right}), our maximum learning rate adjustment facilitated more efficient convergence during training, achieving 31\% FLOPs reduction compared to 26\% reduction observed without learning rate adjustment. 
These results demonstrate the effectiveness of our progressive training approach and show that appropriate learning rate adjustments can further enhance performance\footnote{
Moreover, as shown in Appendix~\ref{appendix:post_training} the benefits of our approach persist under post-training setups (SFT+DPO), confirming its robustness for practical applications.
}.

\section{The Consistency across Model Family}
In this section, we investigate the consistency of model behaviors across different sizes in our progressively trained model family.
We first discuss the advantages of maintaining consistent behaviors across the model family (Section~\ref{subsec:consistency-significance}).
We then present empirical analyses of consistency through probability distribution and speculative decoding. 
Our findings show that models trained using the progressive training approach exhibit higher consistency compared to independently trained models.

\subsection{The Potential Effectiveness of Consistency across Model Family}\label{subsec:consistency-significance}

Ensuring consistent behavior across a model family offers several practical benefits.
First, it simplifies deployment flexibility by ensuring that switching between different parameter sizes does not lead to large, sudden shifts in model outputs or user experience.
This is particularly valuable when developers need to adapt a model size to varying computational budgets or runtime constraints, as they can easily scale up or down (e.g., from a smaller to a larger model) without retraining users or extensively revalidating system performance~\citep{Megha2020compati, echterhoff-etal-2024-muscle}.

Second, such consistency enables more efficient incremental improvements or patches across the entire model family.
For instance, if developers collect preference data or build a reward model based on the outputs of one model (e.g., by annotating the specific texts that this member tends to generate), they typically face a distribution mismatch when applying those artifacts to other models.
In a consistent model family, however, differences in generation patterns are small enough that examples or learned preferences remain effective for other models~\citep{guo2024direct,zhou2024wpo,tajwar2024preference}.
Likewise, implementations such as safety or content filters calibrated for one model will likely transfer to others with only minimal adjustments~\citep{inan2023llamaguard}.

Third, consistency across models of different sizes benefits speculative decoding~\citep{leviathan2023fast}, a technique that accelerates inference by allowing a smaller model to generate ``draft'' outputs, which the larger model then either accepts or refines.
When the smaller and larger models produce similar probability distributions, the larger model is more likely to accept the drafts, reducing the frequency of rejections and subsequent regenerations~\citep{zhou2024distillspec}.
Consequently, consistent behavior across a range of model sizes further contributes to more efficient and user-friendly deployments in real-world applications.

\begin{table}[t]
    \centering
    \small
    \begin{tabular}{cc|c}
        \toprule
        Model Pair & Training Approach &  $D_{\textrm{KL}}(P_{X_{i}} || P_{X_{i+1}})$   \\
        \midrule
        \multirow{3}{*}{1B→2B} 
            & \independent{} & 0.2821 \\
            & \progressive{} & \textbf{0.2162} \\
            & \progressive{}\texttt{+LR adjustment} & 0.2538 \\
        \midrule
        \multirow{3}{*}{2B→4B} 
            & \independent{} & 0.3087 \\
            & \progressive{} & \textbf{0.2265} \\
            & \progressive{}\texttt{+LR adjustment} & 0.2542 \\
        \midrule
        \multirow{3}{*}{4B→8B} 
            & \independent{} & 0.4584 \\
            & \progressive{} & \textbf{0.3378} \\
            & \progressive{}\texttt{+LR adjustment} & 0.3518 \\
        \bottomrule
    \end{tabular}
    \caption{
    KL divergence between adjacent model sizes in the different model families. Lower values indicate greater consistency. Results are shown for models trained from scratch (\independent{}), models built through progressive training (\progressive{}), and models built with maximum learning rate adjustment (\progressive{}\texttt{+LR adjustment}).
    }
    \label{tab:kl_divergence}
\end{table}

\subsection{Probability Distribution Consistency across Model Family}\label{subsec:prob-consistency}

The consistency of the underlying probability distributions across the model family offers deeper insights into how knowledge and behaviors are propagated. 
Such consistency indicates that models rely on similar internal mechanisms when generating text.

\paragraph{Experimental Setup.}

To evaluate the probability distribution consistency across our model family, we measured the KL divergence between adjacent model sizes.
KL divergence quantifies how one probability distribution differs from another reference distribution, with lower values indicating greater similarity.
For this analysis, we used 10,000 examples from the FineWeb-Edu validation dataset.
We calculated the KL divergence $D_{\textrm{KL}}(P_{X_{i}} \parallel P_{X_{i+1}})$ between adjacent models by examining their next-token prediction distributions, where $P_{X_i}$ represents the probability distribution over the vocabulary given by model of size $X_i$\footnote{
The detailed calculation method for KL divergence is described in Appendix~\ref{appendix:kl_comp}．
}. 
We conducted this analysis for three different model families:
(i) Models trained independently from scratch, (ii) Models built through progressive training, (iii) Models built through progressive training with maximum learning rate adjustment.

\paragraph{Results.}
Table~\ref{tab:kl_divergence} presents the KL divergence between adjacent model sizes for our three different model families. 
We found that, across all model pairs, models trained via progressive training demonstrate substantially lower KL divergence compared to independently trained models, indicating higher consistency in their probability distributions. 
This trend is consistent across all adjacent model pairs (1B→2B, 2B→4B, 4B→8B).

This finding provides evidence that our progressive training produces a more coherent model family with consistent internal behavior.
The lower KL divergence suggests that progressively trained models share similar probability distributions and decision-making processes, offering practical advantages for techniques like speculative decoding, where alignment between smaller and larger models is crucial.

\subsection{Validating Consistency Benefits through Speculative Decoding}
We empirically assess whether this consistency translates into practical benefits for speculative decoding, as discussed in Section~\ref{subsec:consistency-significance}.

\paragraph{Experimental Setup.}
In speculative decoding, a smaller ``draft'' model proposes multiple tokens that a large ``generator'' model subsequently accepts or rejects.
For our experiments, we fixed the 8B model by Progressive training as the generator and paired it with 4B draft models trained using either Progressive or Independent training.
Since the effectiveness of speculative decoding depends on how well the draft model's predictions align with the generator's distribution, this setup validates whether the distributional consistency yields practical benefits.
Each draft model was configured to propose 8 speculative tokens.
To evaluate speculative decoding performance, we measured three key metrics on 1,000 prompts sampled from the FineWeb-Edu validation dataset.
First, we computed the KL divergence $D_{\textrm{KL}}(P_{X_{\textrm{draft}}} \parallel P_{X_{\textrm{generator}}})$ between the draft and generator models' output distributions.
Second, we tracked the acceptance rate, which is the percentage of draft tokens accepted by the generator model.
This metric directly reflects the efficiency of speculative decoding.
Finally, we recorded the average generation time across all prompts to quantify the practical speedup.

\paragraph{Results.}
\begin{table}[t]
    \centering
    \small
    \begin{tabular}{cc|ccc}
        \toprule
        Draft Model & Generator Model & $D_{\textrm{KL}}(P_{\text{draft}} \parallel P_{\text{generator}})$ & Acceptance & Generation \\
        (4B) & (8B) & & Rate (\%) & Time (s) \\
        \midrule
        Progressive & Progressive & \textbf{0.3378} & \textbf{93.20} & \textbf{7.02} \\
        Independent & Progressive & 0.4281 & 87.20 & 8.24 \\
        \bottomrule
    \end{tabular}
    \caption{Speculative decoding performance with different draft-generator configurations. Progressive draft models demonstrate lower KL divergence with the generator, resulting in higher acceptance rates and faster generation times.}
    \label{tab:speculative_decoding}
\end{table}

Table~\ref{tab:speculative_decoding} presents the speculative decoding performance with different draft-generator configurations.
The results demonstrate the practical advantages of employing a draft model constructed through Progressive Training. 
The pairing of a 4B Progressive draft model with an 8B Progressive generator model (Prog-Prog) exhibited superior performance, underpinned by greater consistency between the two models. 
This enhanced consistency is quantified by a lower KL divergence for the Prog-Prog pair, compared to using an independently trained model as a drafter (Inde-Prog). 
 This improved alignment in output distributions directly translated to more effective speculative decoding: the Prog-Prog configuration achieved a higher acceptance rate of 93.20\%, as opposed to 87.2\% for the Inde-Prog setup. Consequently, this led to a reduction in inference time, with the Prog-Prog pair completing generation approximately 14.8\% faster. 
 These findings experimentally demonstrate that the consistency fostered by Progressive Training offers practical benefits.
 The lower KL divergence between progressively trained models leads to more effective draft proposals in speculative decoding, thereby improving acceptance rates and reducing inference latency.

\section{Related Work}

\paragraph{Model Family.}
Offering models in multiple sizes has become a standard practice in language model development to address diverse computational requirements. 
Major research labs have released model families with multiple parameter sizes: Llama~\citep{touvron2023llama, grattafiori2024llama3herdmodels}, Qwen~\citep{yang2024qwen2technicalreport}, Gemma~\citep{team2024gemma}, and others. 
Scaling laws~\citep{kaplan2020scaling, hoffmann2022chinchilla} provide the theoretical foundation for these families, establishing relationships between model size, data, compute, and performance. 
Recent research highlights the complementary roles of models of different sizes~\citep{wang2024comprehensive}: while large models excel in zero/few-shot generalization, smaller models offer advantages for latency-sensitive applications, edge deployments, domain-specific tasks, and privacy-sensitive contexts. 
This functional differentiation underscores the importance of efficient methods for constructing a coherent model family that maintains consistent capabilities across different parameter sizes.
However, the conventional approach of training each model size independently incurs substantial computational costs that scale additively with the number of models in the family. 
Our work addresses this inefficiency by proposing progressive training that significantly reduces the total computation required to construct a complete and coherent model family.

\paragraph{Model Expansion.}
Model expansion has emerged as an approach to reduce computational costs when training large-scale models. 
These methods leverage parameters of smaller, pre-trained models to initialize larger ones, accelerating training.
The bert2BERT~\citep{chen-etal-2022-bert2bert} adapted function-preserving methods to Transformers, reducing pre-training costs by approximately 45\% for BERT models. 
LEMON~\citep{wang2024lemon} refined this approach with an optimized learning rate scheduler for larger models. 
For scaling LLMs, \citet{du2024stacking} identified depthwise stacking as particularly effective, achieving 54.6\% speedup for 7B-parameter models. 
Alternative approaches combine knowledge distillation with model expansion. ~\citet{qin-etal-2022-knowledge} introduced Knowledge Inheritance (KI), using knowledge distillation during pre-training to achieve approximately 27\% computational cost reduction. 
However, these methods typically focus on the efficient training of a single, final model rather than the training of an entire model family.
For example, many methods demonstrate the effectiveness of the expansion in a single-shot manner, i.e., applying model expansion only once.
In addition, these methods do not take the performance of intermediate model(s) into account.
In contrast, we propose progressive training and demonstrate that the off-the-shelf model expansion methods can be used to efficiently train a model family.
The progressive training can incorporate emerging novel model expansion techniques, further enhancing efficiency.

\section{Conclusion}
In this paper, we propose an efficient approach for constructing a model family via progressive training, wherein smaller models are incrementally expanded to larger sizes. 
Through comprehensive experiments on a model family ranging from 1B to 8B parameters, we demonstrated that our method significantly reduces the total computational cost required to construct a complete model family by approximately 25\% compared to training each model independently.
Our method, particularly when combined with maximum learning rate adjustments tailored to each model size, not only matches but exceeds the performance of independently trained models.
Furthermore, models built through our progressive approach demonstrate greater consistency across different sizes, as evidenced by lower KL divergence between probability distributions.

This work offers researchers and practitioners a computationally efficient approach to constructing a high-performing, coherent model family, offering particular value in resource-constrained environments where training multiple models at different scales is desirable.

\clearpage
\section*{Acknowledgements}
This work was supported by the JST Moonshot R\&D Grant Number JPMJMS2011-35 (fundamental research).


\begin{thebibliography}{42}
\providecommand{\natexlab}[1]{#1}
\providecommand{\url}[1]{\texttt{#1}}
\expandafter\ifx\csname urlstyle\endcsname\relax
  \providecommand{\doi}[1]{doi: #1}\else
  \providecommand{\doi}{doi: \begingroup \urlstyle{rm}\Url}\fi

\bibitem[Abdin et~al.(2024)Abdin, Aneja, Awadalla, Awadallah, Awan, Bach, Bahree, Bakhtiari, Bao, Behl, et~al.]{abdin2024phi}
Marah Abdin, Jyoti Aneja, Hany Awadalla, Ahmed Awadallah, Ammar~Ahmad Awan, Nguyen Bach, Amit Bahree, Arash Bakhtiari, Jianmin Bao, Harkirat Behl, et~al.
\newblock Phi-3 technical report: A highly capable language model locally on your phone.
\newblock \emph{arXiv preprint arXiv:2404.14219}, 2024.

\bibitem[AI@Meta(2024)]{grattafiori2024llama3herdmodels}
AI@Meta.
\newblock The llama 3 herd of models, 2024.
\newblock URL \url{https://arxiv.org/abs/2407.21783}.

\bibitem[Bisk et~al.(2020)Bisk, Zellers, bras, Gao, and Yejin]{piqa}
Yonatan Bisk, Rowan Zellers, Ronan bras, Jianfeng Gao, and Choi Yejin.
\newblock Piqa: Reasoning about physical commonsense in natural language.
\newblock \emph{Proceedings of the AAAI Conference on Artificial Intelligence}, 34:\penalty0 7432--7439, 04 2020.
\newblock \doi{10.1609/aaai.v34i05.6239}.

\bibitem[Brown et~al.(2020)Brown, Mann, Ryder, Subbiah, Kaplan, Dhariwal, Neelakantan, Shyam, Sastry, Askell, et~al.]{brown2020gpt3}
Tom Brown, Benjamin Mann, Nick Ryder, Melanie Subbiah, Jared~D Kaplan, Prafulla Dhariwal, Arvind Neelakantan, Pranav Shyam, Girish Sastry, Amanda Askell, et~al.
\newblock Language models are few-shot learners.
\newblock \emph{Advances in neural information processing systems}, 33:\penalty0 1877--1901, 2020.

\bibitem[Chen et~al.(2022)Chen, Yin, Shang, Jiang, Qin, Wang, Wang, Chen, Liu, and Liu]{chen-etal-2022-bert2bert}
Cheng Chen, Yichun Yin, Lifeng Shang, Xin Jiang, Yujia Qin, Fengyu Wang, Zhi Wang, Xiao Chen, Zhiyuan Liu, and Qun Liu.
\newblock bert2{BERT}: Towards reusable pretrained language models.
\newblock In Smaranda Muresan, Preslav Nakov, and Aline Villavicencio (eds.), \emph{Proceedings of the 60th Annual Meeting of the Association for Computational Linguistics (Volume 1: Long Papers)}, pp.\  2134--2148, Dublin, Ireland, May 2022. Association for Computational Linguistics.
\newblock \doi{10.18653/v1/2022.acl-long.151}.
\newblock URL \url{https://aclanthology.org/2022.acl-long.151/}.

\bibitem[Clark et~al.(2018)Clark, Cowhey, Etzioni, Khot, Sabharwal, Schoenick, and Tafjord]{clark2018think}
Peter Clark, Isaac Cowhey, Oren Etzioni, Tushar Khot, Ashish Sabharwal, Carissa Schoenick, and Oyvind Tafjord.
\newblock Think you have solved question answering? try arc, the ai2 reasoning challenge, 2018.

\bibitem[Cui et~al.(2024)Cui, Yuan, Ding, Yao, He, Zhu, Ni, Xie, Xie, Lin, Liu, and Sun]{cui2024ultrafeedback}
Ganqu Cui, Lifan Yuan, Ning Ding, Guanming Yao, Bingxiang He, Wei Zhu, Yuan Ni, Guotong Xie, Ruobing Xie, Yankai Lin, Zhiyuan Liu, and Maosong Sun.
\newblock {ULTRAFEEDBACK}: Boosting language models with scaled {AI} feedback.
\newblock In \emph{Forty-first International Conference on Machine Learning}, 2024.
\newblock URL \url{https://openreview.net/forum?id=BOorDpKHiJ}.

\bibitem[Ding et~al.(2023)Ding, Chen, Xu, Qin, Hu, Liu, Sun, and Zhou]{ding-etal-2023-enhancing}
Ning Ding, Yulin Chen, Bokai Xu, Yujia Qin, Shengding Hu, Zhiyuan Liu, Maosong Sun, and Bowen Zhou.
\newblock Enhancing chat language models by scaling high-quality instructional conversations.
\newblock In Houda Bouamor, Juan Pino, and Kalika Bali (eds.), \emph{Proceedings of the 2023 Conference on Empirical Methods in Natural Language Processing}, pp.\  3029--3051, Singapore, December 2023. Association for Computational Linguistics.
\newblock \doi{10.18653/v1/2023.emnlp-main.183}.
\newblock URL \url{https://aclanthology.org/2023.emnlp-main.183/}.

\bibitem[Du et~al.(2024)Du, Luo, Qiu, Huang, Shen, Cheng, Guo, and Fu]{du2024stacking}
Wenyu Du, Tongxu Luo, Zihan Qiu, Zeyu Huang, Yikang Shen, Reynold Cheng, Yike Guo, and Jie Fu.
\newblock Stacking your transformers: A closer look at model growth for efficient {LLM} pre-training.
\newblock In \emph{The Thirty-eighth Annual Conference on Neural Information Processing Systems}, 2024.
\newblock URL \url{https://openreview.net/forum?id=FXJDcriMYH}.

\bibitem[Echterhoff et~al.(2024)Echterhoff, Faghri, Vemulapalli, Hu, Li, Tuzel, and Pouransari]{echterhoff-etal-2024-muscle}
Jessica~Maria Echterhoff, Fartash Faghri, Raviteja Vemulapalli, Ting-Yao Hu, Chun-Liang Li, Oncel Tuzel, and Hadi Pouransari.
\newblock {MUSCLE}: A model update strategy for compatible {LLM} evolution.
\newblock In Yaser Al-Onaizan, Mohit Bansal, and Yun-Nung Chen (eds.), \emph{Findings of the Association for Computational Linguistics: EMNLP 2024}, pp.\  7320--7332, Miami, Florida, USA, November 2024. Association for Computational Linguistics.
\newblock \doi{10.18653/v1/2024.findings-emnlp.430}.
\newblock URL \url{https://aclanthology.org/2024.findings-emnlp.430/}.

\bibitem[GemmaTeam(2024)]{team2024gemma}
GemmaTeam.
\newblock Gemma: Open models based on gemini research and technology.
\newblock \emph{arXiv preprint arXiv:2403.08295}, 2024.

\bibitem[GemmaTeam(2025)]{team2025gemma3}
GemmaTeam.
\newblock Gemma 3 technical report.
\newblock \emph{arXiv preprint arXiv:2503.19786}, 2025.

\bibitem[Guo et~al.(2024)Guo, Zhang, Liu, Liu, Khalman, Llinares, Rame, Mesnard, Zhao, Piot, et~al.]{guo2024direct}
Shangmin Guo, Biao Zhang, Tianlin Liu, Tianqi Liu, Misha Khalman, Felipe Llinares, Alexandre Rame, Thomas Mesnard, Yao Zhao, Bilal Piot, et~al.
\newblock Direct language model alignment from online ai feedback.
\newblock \emph{arXiv preprint arXiv:2402.04792}, 2024.

\bibitem[Hoffmann et~al.(2022)Hoffmann, Borgeaud, Mensch, Buchatskaya, Cai, Rutherford, de~las Casas, Hendricks, Welbl, Clark, Hennigan, Noland, Millican, van~den Driessche, Damoc, Guy, Osindero, Simonyan, Elsen, Vinyals, Rae, and Sifre]{hoffmann2022chinchilla}
Jordan Hoffmann, Sebastian Borgeaud, Arthur Mensch, Elena Buchatskaya, Trevor Cai, Eliza Rutherford, Diego de~las Casas, Lisa~Anne Hendricks, Johannes Welbl, Aidan Clark, Tom Hennigan, Eric Noland, Katherine Millican, George van~den Driessche, Bogdan Damoc, Aurelia Guy, Simon Osindero, Karen Simonyan, Erich Elsen, Oriol Vinyals, Jack~William Rae, and Laurent Sifre.
\newblock An empirical analysis of compute-optimal large language model training.
\newblock In Alice~H. Oh, Alekh Agarwal, Danielle Belgrave, and Kyunghyun Cho (eds.), \emph{Advances in Neural Information Processing Systems}, 2022.
\newblock URL \url{https://openreview.net/forum?id=iBBcRUlOAPR}.

\bibitem[Inan et~al.(2023)Inan, Upasani, Chi, Rungta, Iyer, Mao, Tontchev, Hu, Fuller, Testuggine, et~al.]{inan2023llamaguard}
Hakan Inan, Kartikeya Upasani, Jianfeng Chi, Rashi Rungta, Krithika Iyer, Yuning Mao, Michael Tontchev, Qing Hu, Brian Fuller, Davide Testuggine, et~al.
\newblock Llama guard: {LLM}-based input-output safeguard for {Human-AI} conversations.
\newblock \emph{arXiv preprint arXiv:2312.06674}, 2023.

\bibitem[Kaplan et~al.(2020)Kaplan, McCandlish, Henighan, Brown, Chess, Child, Gray, Radford, Wu, and Amodei]{kaplan2020scaling}
Jared Kaplan, Sam McCandlish, Tom Henighan, Tom~B Brown, Benjamin Chess, Rewon Child, Scott Gray, Alec Radford, Jeffrey Wu, and Dario Amodei.
\newblock Scaling laws for neural language models.
\newblock \emph{arXiv preprint arXiv:2001.08361}, 2020.

\bibitem[Leviathan et~al.(2023)Leviathan, Kalman, and Matias]{leviathan2023fast}
Yaniv Leviathan, Matan Kalman, and Yossi Matias.
\newblock Fast inference from transformers via speculative decoding.
\newblock In \emph{International Conference on Machine Learning}, pp.\  19274--19286. PMLR, 2023.

\bibitem[Meng et~al.(2024)Meng, Xia, and Chen]{meng2024simpo}
Yu~Meng, Mengzhou Xia, and Danqi Chen.
\newblock Sim{PO}: Simple preference optimization with a reference-free reward.
\newblock In \emph{The Thirty-eighth Annual Conference on Neural Information Processing Systems}, 2024.
\newblock URL \url{https://openreview.net/forum?id=3Tzcot1LKb}.

\bibitem[Merity et~al.(2017)Merity, Xiong, Bradbury, and Socher]{merity2017pointer}
Stephen Merity, Caiming Xiong, James Bradbury, and Richard Socher.
\newblock Pointer sentinel mixture models.
\newblock In \emph{International Conference on Learning Representations}, 2017.
\newblock URL \url{https://openreview.net/forum?id=Byj72udxe}.

\bibitem[Mihaylov et~al.(2018)Mihaylov, Clark, Khot, and Sabharwal]{mihaylov-etal-2018-suit}
Todor Mihaylov, Peter Clark, Tushar Khot, and Ashish Sabharwal.
\newblock Can a suit of armor conduct electricity? a new dataset for open book question answering.
\newblock In Ellen Riloff, David Chiang, Julia Hockenmaier, and Jun{'}ichi Tsujii (eds.), \emph{Proceedings of the 2018 Conference on Empirical Methods in Natural Language Processing}, pp.\  2381--2391, Brussels, Belgium, October-November 2018. Association for Computational Linguistics.
\newblock \doi{10.18653/v1/D18-1260}.
\newblock URL \url{https://aclanthology.org/D18-1260}.

\bibitem[Paperno et~al.(2016)Paperno, Kruszewski, Lazaridou, Pham, Bernardi, Pezzelle, Baroni, Boleda, and Fern{\'a}ndez]{paperno-etal-2016-lambada}
Denis Paperno, Germ{\'a}n Kruszewski, Angeliki Lazaridou, Ngoc~Quan Pham, Raffaella Bernardi, Sandro Pezzelle, Marco Baroni, Gemma Boleda, and Raquel Fern{\'a}ndez.
\newblock The {LAMBADA} dataset: Word prediction requiring a broad discourse context.
\newblock In Katrin Erk and Noah~A. Smith (eds.), \emph{Proceedings of the 54th Annual Meeting of the Association for Computational Linguistics (Volume 1: Long Papers)}, pp.\  1525--1534, Berlin, Germany, August 2016. Association for Computational Linguistics.
\newblock \doi{10.18653/v1/P16-1144}.
\newblock URL \url{https://aclanthology.org/P16-1144}.

\bibitem[Penedo et~al.(2024)Penedo, Kydl{\'\i}{\v{c}}ek, allal, Lozhkov, Mitchell, Raffel, Werra, and Wolf]{penedo2024fineweb}
Guilherme Penedo, Hynek Kydl{\'\i}{\v{c}}ek, Loubna~Ben allal, Anton Lozhkov, Margaret Mitchell, Colin Raffel, Leandro~Von Werra, and Thomas Wolf.
\newblock The {FineWeb} datasets: Decanting the {Web} for the finest text data at scale.
\newblock In \emph{The Thirty-eight Conference on Neural Information Processing Systems Datasets and Benchmarks Track}, 2024.
\newblock URL \url{https://openreview.net/forum?id=n6SCkn2QaG}.

\bibitem[Qin et~al.(2022)Qin, Lin, Yi, Zhang, Han, Zhang, Su, Liu, Li, Sun, and Zhou]{qin-etal-2022-knowledge}
Yujia Qin, Yankai Lin, Jing Yi, Jiajie Zhang, Xu~Han, Zhengyan Zhang, Yusheng Su, Zhiyuan Liu, Peng Li, Maosong Sun, and Jie Zhou.
\newblock Knowledge inheritance for pre-trained language models.
\newblock In Marine Carpuat, Marie-Catherine de~Marneffe, and Ivan~Vladimir Meza~Ruiz (eds.), \emph{Proceedings of the 2022 Conference of the North American Chapter of the Association for Computational Linguistics: Human Language Technologies}, pp.\  3921--3937, Seattle, United States, July 2022. Association for Computational Linguistics.
\newblock \doi{10.18653/v1/2022.naacl-main.288}.
\newblock URL \url{https://aclanthology.org/2022.naacl-main.288/}.

\bibitem[Radford et~al.(2019)Radford, Wu, Child, Luan, Amodei, and Sutskever]{radford2019language}
Alec Radford, Jeff Wu, Rewon Child, David Luan, Dario Amodei, and Ilya Sutskever.
\newblock Language models are unsupervised multitask learners, 2019.

\bibitem[Rafailov et~al.(2023)Rafailov, Sharma, Mitchell, Manning, Ermon, and Finn]{rafailov2023direct}
Rafael Rafailov, Archit Sharma, Eric Mitchell, Christopher~D Manning, Stefano Ermon, and Chelsea Finn.
\newblock Direct preference optimization: Your language model is secretly a reward model.
\newblock In \emph{Thirty-seventh Conference on Neural Information Processing Systems}, 2023.
\newblock URL \url{https://openreview.net/forum?id=HPuSIXJaa9}.

\bibitem[Sakaguchi et~al.(2021)Sakaguchi, Bras, Bhagavatula, and Choi]{sakaguchi2021winogrande}
Keisuke Sakaguchi, Ronan~Le Bras, Chandra Bhagavatula, and Yejin Choi.
\newblock Winogrande: An adversarial winograd schema challenge at scale.
\newblock \emph{Communications of the ACM}, 64\penalty0 (9):\penalty0 99--106, 2021.

\bibitem[Sardana et~al.(2024)Sardana, Portes, Doubov, and Frankle]{sardana2024beyond}
Nikhil Sardana, Jacob Portes, Sasha Doubov, and Jonathan Frankle.
\newblock Beyond chinchilla-optimal: Accounting for inference in language model scaling laws.
\newblock In \emph{Forty-first International Conference on Machine Learning}, 2024.
\newblock URL \url{https://openreview.net/forum?id=0bmXrtTDUu}.

\bibitem[Srivastava et~al.(2020)Srivastava, Nushi, Kamar, Shah, and Horvitz]{Megha2020compati}
Megha Srivastava, Besmira Nushi, Ece Kamar, Shital Shah, and Eric Horvitz.
\newblock An empirical analysis of backward compatibility in machine learning systems.
\newblock In \emph{Proceedings of the 26th ACM SIGKDD International Conference on Knowledge Discovery \& Data Mining}, KDD '20, pp.\  3272–3280, New York, NY, USA, 2020. Association for Computing Machinery.
\newblock ISBN 9781450379984.
\newblock \doi{10.1145/3394486.3403379}.
\newblock URL \url{https://doi.org/10.1145/3394486.3403379}.

\bibitem[Tajwar et~al.(2024)Tajwar, Singh, Sharma, Rafailov, Schneider, Xie, Ermon, Finn, and Kumar]{tajwar2024preference}
Fahim Tajwar, Anikait Singh, Archit Sharma, Rafael Rafailov, Jeff Schneider, Tengyang Xie, Stefano Ermon, Chelsea Finn, and Aviral Kumar.
\newblock Preference fine-tuning of {LLMs} should leverage suboptimal, on-policy data.
\newblock In \emph{Proceedings of the 41st International Conference on Machine Learning}, 2024.

\bibitem[Takase et~al.(2023)Takase, Kiyono, Kobayashi, and Suzuki]{takase2023spike}
Sho Takase, Shun Kiyono, Sosuke Kobayashi, and Jun Suzuki.
\newblock Spike no more: Stabilizing the pre-training of large language models.
\newblock \emph{arXiv preprint arXiv:2312.16903}, 2023.

\bibitem[Touvron et~al.(2023)Touvron, Lavril, Izacard, Martinet, Lachaux, Lacroix, Rozi{\`e}re, Goyal, Hambro, Azhar, et~al.]{touvron2023llama}
Hugo Touvron, Thibaut Lavril, Gautier Izacard, Xavier Martinet, Marie-Anne Lachaux, Timoth{\'e}e Lacroix, Baptiste Rozi{\`e}re, Naman Goyal, Eric Hambro, Faisal Azhar, et~al.
\newblock Llama: Open and efficient foundation language models.
\newblock \emph{arXiv preprint arXiv:2302.13971}, 2023.

\bibitem[Vaswani et~al.(2017)Vaswani, Shazeer, Parmar, Uszkoreit, Jones, Gomez, Kaiser, and Polosukhin]{vaswani2017attention}
Ashish Vaswani, Noam Shazeer, Niki Parmar, Jakob Uszkoreit, Llion Jones, Aidan~N Gomez, {\L}ukasz Kaiser, and Illia Polosukhin.
\newblock Attention is all you need.
\newblock \emph{Advances in neural information processing systems}, 30, 2017.

\bibitem[Wang et~al.(2024{\natexlab{a}})Wang, Zhang, Zhang, Wu, Mo, Lu, Wang, Li, Xu, Tang, et~al.]{wang2024comprehensive}
Fali Wang, Zhiwei Zhang, Xianren Zhang, Zongyu Wu, Tzuhao Mo, Qiuhao Lu, Wanjing Wang, Rui Li, Junjie Xu, Xianfeng Tang, et~al.
\newblock A comprehensive survey of small language models in the era of large language models: Techniques, enhancements, applications, collaboration with {LLMs}, and trustworthiness.
\newblock \emph{arXiv preprint arXiv:2411.03350}, 2024{\natexlab{a}}.

\bibitem[Wang et~al.(2024{\natexlab{b}})Wang, Su, Lu, Xie, Liu, Yuan, Lin, Sun, and Yang]{wang2024lemon}
Yite Wang, Jiahao Su, Hanlin Lu, Cong Xie, Tianyi Liu, Jianbo Yuan, Haibin Lin, Ruoyu Sun, and Hongxia Yang.
\newblock {LEMON}: Lossless model expansion.
\newblock In \emph{The Twelfth International Conference on Learning Representations}, 2024{\natexlab{b}}.
\newblock URL \url{https://openreview.net/forum?id=3Vw7DQqq7U}.

\bibitem[Wei et~al.(2022)Wei, Tay, Bommasani, Raffel, Zoph, Borgeaud, Yogatama, Bosma, Zhou, Metzler, Chi, Hashimoto, Vinyals, Liang, Dean, and Fedus]{wei2022emergent}
Jason Wei, Yi~Tay, Rishi Bommasani, Colin Raffel, Barret Zoph, Sebastian Borgeaud, Dani Yogatama, Maarten Bosma, Denny Zhou, Donald Metzler, Ed~H. Chi, Tatsunori Hashimoto, Oriol Vinyals, Percy Liang, Jeff Dean, and William Fedus.
\newblock Emergent abilities of large language models.
\newblock \emph{Transactions on Machine Learning Research}, 2022.
\newblock ISSN 2835-8856.
\newblock URL \url{https://openreview.net/forum?id=yzkSU5zdwD}.
\newblock Survey Certification.

\bibitem[Wortsman et~al.(2024)Wortsman, Liu, Xiao, Everett, Alemi, Adlam, Co-Reyes, Gur, Kumar, Novak, Pennington, Sohl-Dickstein, Xu, Lee, Gilmer, and Kornblith]{wortsman2024smallscale}
Mitchell Wortsman, Peter~J Liu, Lechao Xiao, Katie~E Everett, Alexander~A Alemi, Ben Adlam, John~D Co-Reyes, Izzeddin Gur, Abhishek Kumar, Roman Novak, Jeffrey Pennington, Jascha Sohl-Dickstein, Kelvin Xu, Jaehoon Lee, Justin Gilmer, and Simon Kornblith.
\newblock Small-scale proxies for large-scale transformer training instabilities.
\newblock In \emph{The Twelfth International Conference on Learning Representations}, 2024.
\newblock URL \url{https://openreview.net/forum?id=d8w0pmvXbZ}.

\bibitem[Yang et~al.(2024{\natexlab{a}})Yang, Yang, Hui, Zheng, Yu, Zhou, Li, Li, Liu, Huang, Dong, Wei, Lin, Tang, Wang, Yang, Tu, Zhang, Ma, Yang, Xu, Zhou, Bai, He, Lin, Dang, Lu, Chen, Yang, Li, Xue, Ni, Zhang, Wang, Peng, Men, Gao, Lin, Wang, Bai, Tan, Zhu, Li, Liu, Ge, Deng, Zhou, Ren, Zhang, Wei, Ren, Liu, Fan, Yao, Zhang, Wan, Chu, Liu, Cui, Zhang, Guo, and Fan]{yang2024qwen2technicalreport}
An~Yang, Baosong Yang, Binyuan Hui, Bo~Zheng, Bowen Yu, Chang Zhou, Chengpeng Li, Chengyuan Li, Dayiheng Liu, Fei Huang, Guanting Dong, Haoran Wei, Huan Lin, Jialong Tang, Jialin Wang, Jian Yang, Jianhong Tu, Jianwei Zhang, Jianxin Ma, Jianxin Yang, Jin Xu, Jingren Zhou, Jinze Bai, Jinzheng He, Junyang Lin, Kai Dang, Keming Lu, Keqin Chen, Kexin Yang, Mei Li, Mingfeng Xue, Na~Ni, Pei Zhang, Peng Wang, Ru~Peng, Rui Men, Ruize Gao, Runji Lin, Shijie Wang, Shuai Bai, Sinan Tan, Tianhang Zhu, Tianhao Li, Tianyu Liu, Wenbin Ge, Xiaodong Deng, Xiaohuan Zhou, Xingzhang Ren, Xinyu Zhang, Xipin Wei, Xuancheng Ren, Xuejing Liu, Yang Fan, Yang Yao, Yichang Zhang, Yu~Wan, Yunfei Chu, Yuqiong Liu, Zeyu Cui, Zhenru Zhang, Zhifang Guo, and Zhihao Fan.
\newblock Qwen2 technical report, 2024{\natexlab{a}}.
\newblock URL \url{https://arxiv.org/abs/2407.10671}.

\bibitem[Yang et~al.(2024{\natexlab{b}})Yang, Yang, Zhang, Hui, Zheng, Yu, Li, Liu, Huang, Wei, et~al.]{yang2024qwen2.5}
An~Yang, Baosong Yang, Beichen Zhang, Binyuan Hui, Bo~Zheng, Bowen Yu, Chengyuan Li, Dayiheng Liu, Fei Huang, Haoran Wei, et~al.
\newblock Qwen2. 5 technical report.
\newblock \emph{arXiv preprint arXiv:2412.15115}, 2024{\natexlab{b}}.

\bibitem[Zellers et~al.(2019)Zellers, Holtzman, Bisk, Farhadi, and Choi]{zellers-etal-2019-hellaswag}
Rowan Zellers, Ari Holtzman, Yonatan Bisk, Ali Farhadi, and Yejin Choi.
\newblock {H}ella{S}wag: Can a machine really finish your sentence?
\newblock In Anna Korhonen, David Traum, and Llu{\'\i}s M{\`a}rquez (eds.), \emph{Proceedings of the 57th Annual Meeting of the Association for Computational Linguistics}, pp.\  4791--4800, Florence, Italy, July 2019. Association for Computational Linguistics.
\newblock \doi{10.18653/v1/P19-1472}.
\newblock URL \url{https://aclanthology.org/P19-1472}.

\bibitem[Zheng et~al.(2023)Zheng, Chiang, Sheng, Zhuang, Wu, Zhuang, Lin, Li, Li, Xing, Zhang, Gonzalez, and Stoica]{zheng2023judging}
Lianmin Zheng, Wei-Lin Chiang, Ying Sheng, Siyuan Zhuang, Zhanghao Wu, Yonghao Zhuang, Zi~Lin, Zhuohan Li, Dacheng Li, Eric Xing, Hao Zhang, Joseph~E. Gonzalez, and Ion Stoica.
\newblock Judging {LLM}-as-a-judge with {MT}-bench and chatbot arena.
\newblock In \emph{Thirty-seventh Conference on Neural Information Processing Systems Datasets and Benchmarks Track}, 2023.
\newblock URL \url{https://openreview.net/forum?id=uccHPGDlao}.

\bibitem[Zhou et~al.(2024{\natexlab{a}})Zhou, Agrawal, Zhang, Indurthi, Zhao, Song, Xu, and Zhu]{zhou2024wpo}
Wenxuan Zhou, Ravi Agrawal, Shujian Zhang, Sathish~Reddy Indurthi, Sanqiang Zhao, Kaiqiang Song, Silei Xu, and Chenguang Zhu.
\newblock {WPO}: Enhancing {RLHF} with weighted preference optimization.
\newblock In \emph{Proceedings of the 2024 Conference on Empirical Methods in Natural Language Processing}, November 2024{\natexlab{a}}.
\newblock \doi{10.18653/v1/2024.emnlp-main.475}.
\newblock URL \url{https://aclanthology.org/2024.emnlp-main.475/}.

\bibitem[Zhou et~al.(2024{\natexlab{b}})Zhou, Lyu, Rawat, Menon, Rostamizadeh, Kumar, Kagy, and Agarwal]{zhou2024distillspec}
Yongchao Zhou, Kaifeng Lyu, Ankit~Singh Rawat, Aditya~Krishna Menon, Afshin Rostamizadeh, Sanjiv Kumar, Jean-Fran{\c{c}}ois Kagy, and Rishabh Agarwal.
\newblock Distillspec: Improving speculative decoding via knowledge distillation.
\newblock In \emph{The Twelfth International Conference on Learning Representations}, 2024{\natexlab{b}}.
\newblock URL \url{https://openreview.net/forum?id=rsY6J3ZaTF}.

\end{thebibliography}

\bibliographystyle{colm2025_conference}

\appendix
\clearpage
\section{Details of Experimental Configurations}\label{appendix:exp_config}

\begin{table}[t]
   \centering
   \small
   \begin{tabular}{ccccc}
       \toprule
       Configuration & 1B & 2B & 4B & 8B \\
       \midrule
       Hidden dimension & 2048 & 2560 & 3200 & 4096 \\
       FFN dimension & 7168 & 8960 & 11200 & 14336 \\
       Layers & 18 & 22 & 27 & 33 \\
       Heads & 16 & 20 & 25 & 32 \\
       Batch size & 960 & 1920 & 3840 & 7680\\
       The number of updates & 40700 & 40700 & 40700 & 40700\\
       learning rate warmup fraction & 0.01 & 0.01 & 0.01 & 0.01 \\
       Learning rate decay style & cosine & cosine & cosine & cosine \\
       Adam $\beta_1$ & 0.9 & 0.9 & 0.9 & 0.9\\
       Adam $\beta_2$ & 0.95 & 0.95 & 0.95 & 0.95\\
       Gradient clipping & 1.0 & 1.0 & 1.0 & 1.0\\
       Weight decay & 0.1 & 0.1 & 0.1 & 0.1\\
       Precision & \texttt{bfloat16} & \texttt{bfloat16} & \texttt{bfloat16} & \texttt{bfloat16} \\
       \bottomrule
   \end{tabular}
   \caption{Experimental configuration for each model size}
   \label{table:model_configs}
\end{table}
In our experiments (Section~\ref{sec:exp_pretraining}), we constructed models with 1B, 2B, 4B, and 8B parameters based on the Llama architecture.
Table~\ref{table:model_configs} shows the specific experimental configurations for each model size.
As the model expands, we gradually increase the hidden dimension, FFN dimension, number of layers, and number of heads.

\section{Data Size for (2x) Chinchilla Law Setting}\label{appendix:chinchilla_setting}
As described in Section~\ref{subsec:data_size}, we determined the number of training tokens for each model size according to \citet{hoffmann2022chinchilla}. 
We report token allocations for each model in Table~\ref{tab:token_allocation_chinchilla} (Chinchilla law setting) and Table~\ref{tab:token_allocation_2x_chinchilla} (2x Chinchilla law setting).

\begin{table}[t]
    \centering
    \small
    \begin{tabular}{ccc}
    \toprule
    Model Size & $T_i^{\textrm{scratch}}$ (\independent{}) & $T_i^{\textrm{prog}}$ (\progressive{}) \\
    \midrule
    1B & 20B & 20B \\
    2B & 40B & 30B \\
    4B & 80B & 60B \\
    8B & 160B & 120B \\
    \bottomrule
    \end{tabular}
    \caption{Training token allocation for \independent{} and \progressive{} the Chinchilla law setting.}
    \label{tab:token_allocation_chinchilla}
\end{table}

\begin{table}[t]
    \centering
    \begin{tabular}{ccc}
    \toprule
    Model Size & $T_i^{\textrm{scratch}}$ (\independent{}) & $T_i^{\textrm{prog}}$ (\progressive{}) \\
    \midrule
    1B & 40B & 40B \\
    2B & 80B & 60B \\
    4B & 160B & 120B \\
    8B & 320B & 240B \\
    \bottomrule
    \end{tabular}
    \caption{Training token allocation for \independent{} and \progressive{} the 2x Chinchilla law setting.}
    \label{tab:token_allocation_2x_chinchilla}
\end{table}

\section{8B Model Loss Visualization}\label{appendix:8b_loss}
\begin{figure}[t]
    \centering
    \includegraphics[width=\columnwidth]{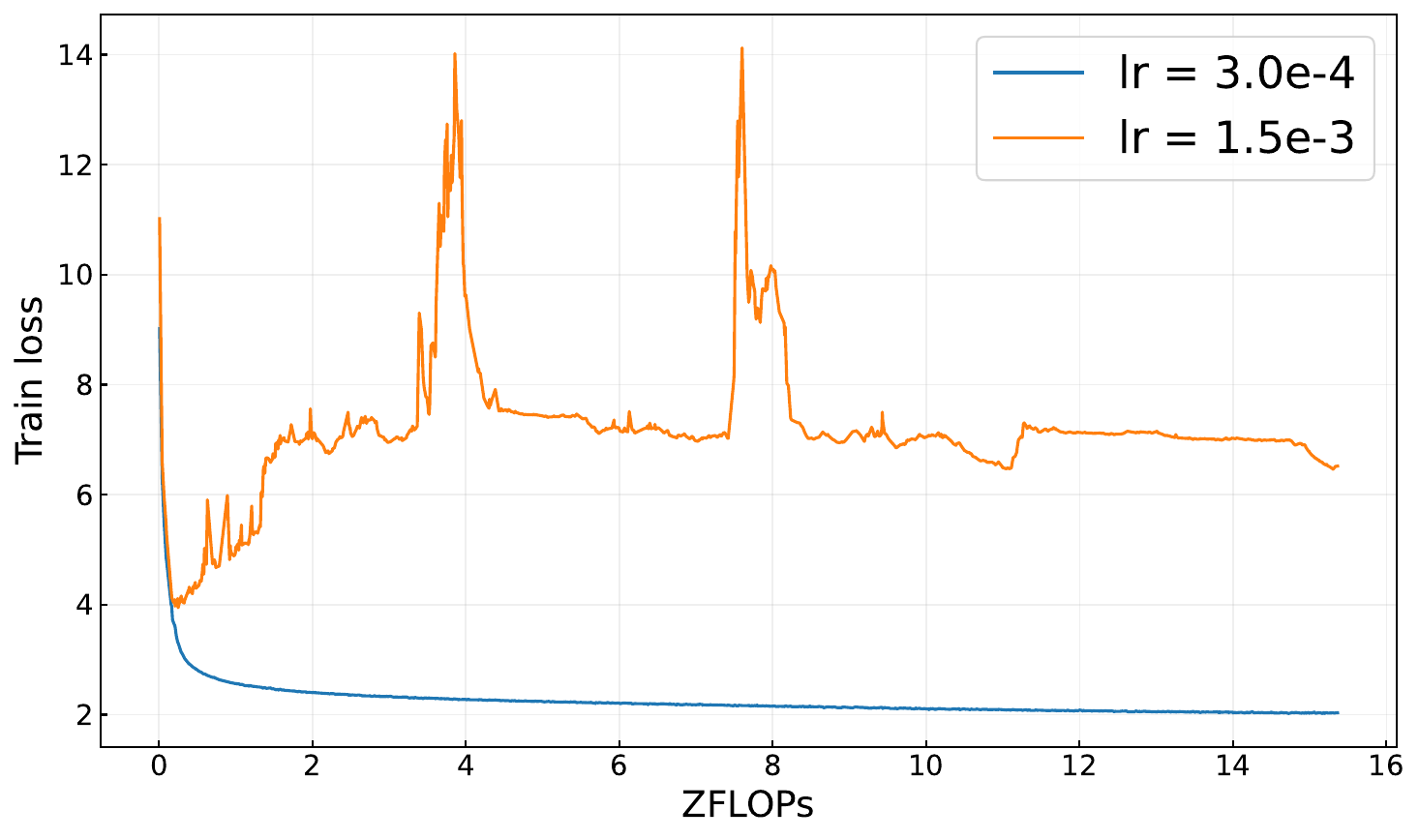}
    \caption{Training loss curves comparing 8B models trained from scratch with different learning rates. The model trained with a learning rate of $3.0 \times 10^{-4}$ (blue line) shows stable training, while the model trained with a higher learning rate of $1.5 \times 10^{-3}$ (orange line) exhibits severe loss spikes indicating training instability.}
    \label{fig:8b_loss_compare}
\end{figure}
Figure~\ref{fig:8b_loss_compare} illustrates the training loss curves for 8B models trained from scratch with different learning rates: $3.0 \times 10^{-4}$ and $1.5 \times 10^{-3}$. While the lower learning rate of $3.0 \times 10^{-4}$ results in stable training, the higher learning rate of $1.5 \times 10^{-3}$ causes significant loss spikes during training.

\section{Specific Configurations in the Learning Rate Adjustment Strategy}\label{appendix:lr_adjustment}

In our learning rate adjustment strategy (Section~\ref{subsec:learning_rate_adjustment}), we decrease the maximum learning rate as the model size increases.
We apply maximum learning rates of $1.5 \times 10^{-3}$, $1.1 \times 10^{-3}$, $7.0 \times 10^{-4}$, and $3.0 \times 10^{-4}$ for the 1B, 2B, 4B, and 8B models, respectively. 
This gradual reduction accommodates the increasing sensitivity of larger models to high learning rates.

\section{Experiment: Model Family Post-Training}\label{appendix:post_training}
\begin{table}[t]
    \centering
    \small
    \begin{tabular}{cc|cc|c}
        \toprule
        \multirow{3}{*}{Size} & \multirow{3}{*}{Training Approach} & \multicolumn{2}{c|}{} & Adjusted LR \\
        & & Chinchilla & 2x Chinchilla & 2x Chinchilla \\
        & & MT-Bench \textuparrow & MT-Bench \textuparrow & MT-Bench \textuparrow \\
        \midrule
        \multirow{2}{*}{2B} 
            & \independent{} & 1.61 & 2.04 & 2.04 \\
            & \progressive{} & \textbf{2.03} & \textbf{2.46} & \textbf{3.22} \\
        \midrule
        \multirow{2}{*}{4B} 
            & \independent{} & 2.78 & \textbf{3.02} & 3.02 \\
            & \progressive{} & \textbf{2.87} & 3.01 & \textbf{3.63} \\
        \midrule
        \multirow{3}{*}{8B} 
            & \independent{} & \textbf{3.42} & \textbf{3.45} & 3.45 \\
            & \progressive{} & 3.28 & 3.21 & \textbf{3.96} \\
            & \progressive{}\texttt{+Fixed Data} & 3.32 & 3.41 & \textbf{3.98} \\
        \bottomrule
    \end{tabular}
    \caption{
    MT-Bench evaluation results after post-training (SFT+DPO) for models trained with different approaches. We compare models trained from scratch (\independent{}) versus those built using our progressive training approach (\progressive{}) under both fixed maximum learning rate and learning rate adjustment (Adjusted LR) strategies. For the fixed learning rate, results are shown for both the Chinchilla law and 2x Chinchilla law settings. For the models with learning rate adjustments, the maximum learning rates are decreased from $1.5 \times 10^{-3}$ for 1B to $3.0 \times 10^{-4}$ for 8B as the model size increases.
    }
    \label{tab:post_training_results}
\end{table}

This experiment verifies whether the trends observed in pre-training experiments in Section~\ref{sec:exp_pretraining} are also seen in post-training. 
We examine whether models built using progressive training with fixed learning rates can achieve comparable performance to models trained from scratch.
In addition, we confirm that performance improvements observed with maximum learning rate adjustments in pre-training also extend to post-training.

\subsection{Experimental Setup}
We evaluate the effectiveness of progressive training by conducting post-training on the model family constructed in Section~\ref{sec:exp_pretraining}.
We adopted the settings from~\citet{meng2024simpo} for post-training.
Specifically, we performed Supervised Fine-Tuning (SFT) using the UltraChat-200k dataset~\citep{ding-etal-2023-enhancing}, followed by DPO~\citep{rafailov2023direct} using the Ultrachat Feedback dataset~\citep{cui2024ultrafeedback}.

The maximum learning rate was set to $3.0 \times 10^{-5}$ for SFT and $5.0 \times 10^{-7}$ for DPO, standardized across all models.
We evaluated the response quality of each model using MT-Bench~\citep{zheng2023judging} using \texttt{gpt-4-0613} as an evaluator.
For comparison, we prepared models (2B, 4B, 8B) independently trained from scratch and conducted the same post-training.

\subsection{Results}
Table~\ref{tab:post_training_results} presents the MT-Bench evaluation results after post-training across different training approaches.
With fixed learning rates, models trained with progressive training perform comparably to models trained from scratch, with some variation across model sizes.

When applying our maximum learning rate adjustment, progressive training consistently outperforms from-scratch training across all model sizes, with substantial improvements observed.
The models with maximum learning rate adjustment significantly outperform those with fixed maximum learning rates for progressive training.

These results indicate that our progressive training approach, particularly when combined with maximum learning rate adjustment, can deliver both computational efficiency and enhanced model performance after post-training.

\section{KL Divergence Computation}\label{appendix:kl_comp}
To simplify the computation, at each token position, we computed $D_{\textrm{KL}}(P_{X_{i}} \parallel P_{X_{i+1}}) = \sum_{v \in V} P_{X_i}(v) \cdot (\log P_{X_i}(v) - \log P_{X_{i+1}}(v))$, where $V$ is the vocabulary and $P_{X_i}(v)$ is the probability assigned to vocabulary token $v$ by model $X_i$. 
We first averaged these KL divergence values across all token positions within each example, then computed the final KL divergence by averaging across all 10,000 examples from the FineWeb-Edu validation dataset, as described in Section~\ref{subsec:prob-consistency}.

\end{document}